\documentclass[conference]{IEEEtran}
\IEEEoverridecommandlockouts

\usepackage{cite}
\usepackage{amsmath,amssymb,amsfonts}
\usepackage{algorithmic}
\usepackage{graphicx}
\usepackage{textcomp}
\usepackage{xcolor}
\usepackage{multirow}
\usepackage{graphicx}
\usepackage{url}
\usepackage{hyperref}

\hypersetup{
    colorlinks=true,
    linkcolor=black,
    citecolor=black,
    urlcolor=black
}

\def\BibTeX{{\rm B\kern-.05em{\sc i\kern-.025em b}\kern-.08em
    T\kern-.1667em\lower.7ex\hbox{E}\kern-.125emX}}
\begin{document}

\title{Cross-Anatomy Transfer Versus Sparse Interpolation in Digital-Twin-Oriented Aortic Fluid--Structure Interaction Surrogates}

\author{\IEEEauthorblockN{Ali Nourbakhsh}
\IEEEauthorblockA{\textit{Department of Mechanical Engineering} \\
\textit{Isfahan University of Technology}\\
Isfahan, Iran \\
nourbakhsh.a@me.iut.ac.ir}
\and
\IEEEauthorblockN{Mohammad Reza Niroomand\textsuperscript{*}}
\IEEEauthorblockA{\textit{Department of Mechanical Engineering} \\
\textit{Isfahan University of Technology}\\
Isfahan, Iran \\
niroomand@iut.ac.ir}
\and
\IEEEauthorblockN{Erfan Nourbakhsh}
\IEEEauthorblockA{\textit{Artificial Intelligence Department} \\
\textit{University of Isfahan}\\
Isfahan, Iran \\
erfannourbakhsh2001@gmail.com}
\thanks{\textsuperscript{*} Corresponding author.}
}

\maketitle

\begin{abstract}
Surrogate credibility for fluid--structure interaction (FSI) requires distinguishing transfer across independent anatomies from interpolation within an already sampled surface. Four de-identified human aortic models from the Vascular Model Repository were reconstructed into separate lumen and nominal 1.5-mm wall domains and analyzed under matched first-cycle two-way FSI. A geometry-only LightGBM prior, selected by leave-one-anatomy-out development on three anatomies, was zero-shot evaluated on a fourth, then probed with a post-zero-shot sparse field-completion case study over six targets. Zero-shot transfer was poor across all targets. At a five-percent anchor level (203 anchors, 3,852 evaluation nodes), prior-plus-adaptation reached an oscillatory shear index (OSI) $R^2$ of 0.603. However, same-anchor controls tuned only on the three development anatomies were stronger for several outcomes: inverse-distance weighting reached $R^2 = 0.829$ (OSI), 0.617 (peak von Mises stress), 0.676 (mean stress); radial basis function interpolation reached 0.917, 0.714, 0.778. Sparse within-anatomy labels thus support field completion, but this four-anatomy cohort gives no evidence the cross-anatomy prior adds value beyond direct interpolation. We frame this as a first computational stage toward a measurement-linked digital twin: the surrogate/update layer is evaluated here, while larger cohorts, converged FSI, measurable patient-side inputs, and physics-informed learning remain future work, not a claim of a complete clinical twin. Our code, data and computation files are available at
\href{https://github.com/ali-nourbakhsh2005/Aortic-FSI-Sparse-Field-Completion}
{\nolinkurl{https://github.com/ali-nourbakhsh2005/Aortic-FSI-Sparse-Field-Completion}}.
\end{abstract}


\section{Introduction}
Aortic aneurysms combine localized dilation with altered blood-flow organization and wall mechanics. Patient-specific computational models resolve wall shear stress (WSS), recirculation, and wall stress over three-dimensional anatomy, while two-way fluid--structure interaction (FSI) further couples pulsatile flow to a deformable arterial wall \cite{b1,b2,b4,b5}, informative but costly to prepare and solve repeatedly.

Surrogate modeling is thus attractive across computational fluid mechanics \cite{b22}, though recent cardiovascular work shows ``fast prediction'' spans very different validation problems: deep networks for velocity/pressure/WSS on shape-augmented CFD cohorts \cite{b9}; POD-based reduced-order modeling with ML for vascular WSS \cite{b10}; POD--LSTM and CNN--LSTM FSI surrogates \cite{b11}; POD+ML for aneurysmal hemodynamics \cite{b12}; and physics-constrained graph learning on 105 semi-idealized geometries \cite{b13}, with recent operator-learning work stressing out-of-distribution testing and sparse parameter inference \cite{b14,b15}. These motivate rapid full-field prediction, but make the unit of generalization, time point, parameter, synthetic geometry, or independent anatomy, central to interpretation.

This is acute for mesh-based FSI fields: thousands of spatially correlated nodes from one anatomy are not thousands of independent patients, so random node splits can estimate interpolation within a solved surface while overstating transfer to an unseen anatomy \cite{b6}. A second ambiguity arises once sparse labels from a new anatomy are introduced: improved performance may reflect simple interpolation of a smooth field, not transfer from a multi-anatomy prior. Any ``personalization'' claim should be checked against a control using the same sparse labels but no prior anatomy.

Digital-twin terminology needs similar care. Earlier cardiovascular roadmaps emphasize personalization and validation \cite{b19}, while a recent circulatory-system review distinguishes static digital models from twins that update via real-world data assimilation \cite{b16}; recent implementations estimate individualized physiology from wearable/imaging data \cite{b15,b17}. We use digital-twin-oriented narrowly: this work develops and stress-tests the surrogate/update layer that could feed a future cardiovascular digital twin, without claiming a complete synchronized twin.

Accordingly, we address four questions: RQ1, cross-anatomy performance of a geometry-only surrogate with no target labels from the new anatomy? RQ2, how does non-anchor reconstruction change as target-anatomy anchors increase from 1\% to 10\%? RQ3, does prior-plus-adaptation outperform same-anchor interpolation tuned without access to the target anatomy? RQ4, do development-derived empirical residual bands or local geometry-shift scores diagnostically associate with reconstruction error on the separate anatomy? The contribution is thus an evaluation framework and falsifiable baseline comparison, not a new learning algorithm.

\section{Materials and Methods}

\subsection{Study Design and Evaluation Hierarchy}
Four computational aortic anatomies were analyzed: 8,191, 6,800, 3,203, and 4,055 inner-wall nodes (22,249 total). The first three formed the development set; the fourth was excluded from LightGBM model selection for the zero-shot experiment, though its files were already mounted for preprocessing, target-excluded selection, not target-file blinding. Geometry features, target transforms, and candidate selection were fixed before zero-shot metrics were computed; the implementation then applied a hard-coded 450-tree final refit before A4 scoring, an override audited in the Supplement: re-fitting with each candidate's native tree count shifts some values but leaves weak transfer as the conclusion, with negligible effect after 5\% adaptation. Because A4 was later inspected for secondary analyses formalized only after the zero-shot result, its sparse-calibration results are a post-zero-shot case study, not a locked external test. Fig.~\ref{fig2} summarizes this hierarchy.

\subsection{Repository Source, Geometry Reconstruction, and Data Governance}
All source vascular models came from the public Vascular Model Repository (VMR), which disseminates de-identified cardiovascular imaging/model data for research use \cite{b3}. Only publicly released model files were accessed; no direct identifiers, participant contact, intervention, or new clinical data collection occurred. VMR permits research/development use under its copyright and acknowledgement conditions; original clinical acquisition, consent, and de-identification remain with the contributing source studies, and this work makes no new IRB claim. The original VMR-to-study label mapping was not preserved in local filenames, so we do not infer which A1--A4 label corresponds to a given VMR identifier without documentary evidence.

\begin{figure}[htbp]
\centering
\includegraphics[width=\linewidth]{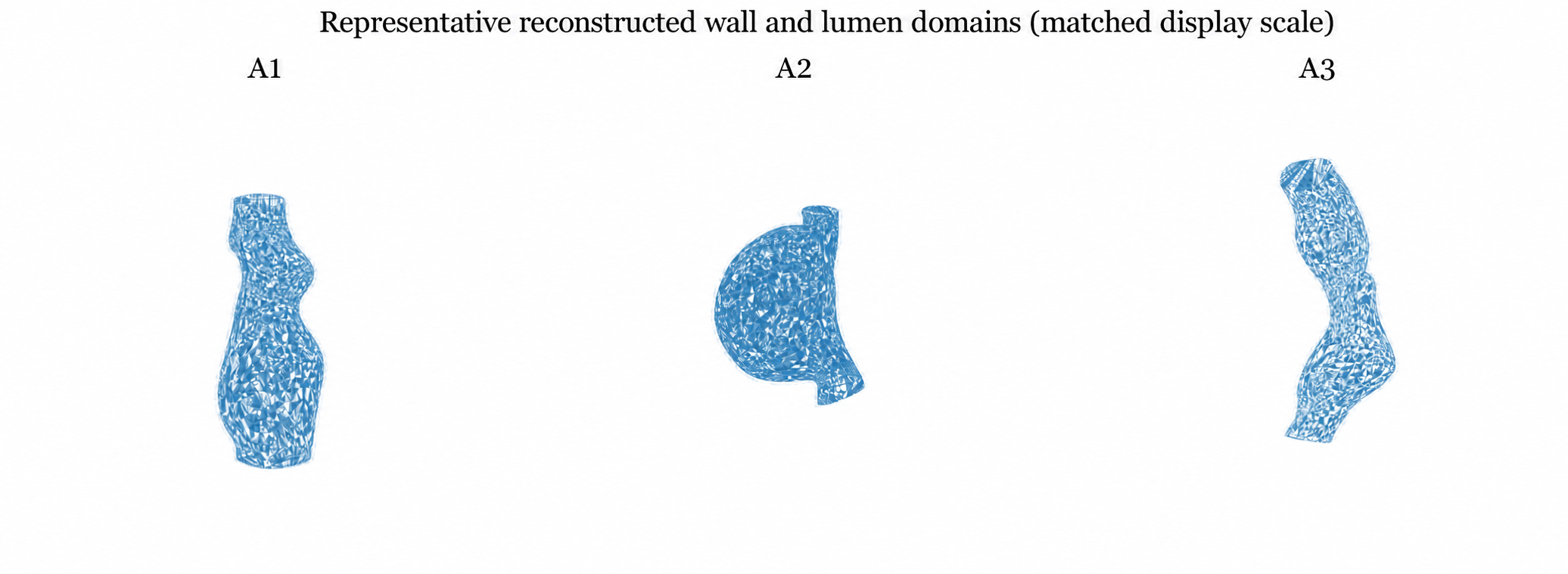}
\caption{Reconstructed geometries for development anatomies A1--A3, matched display scale after principal-axis alignment. Opaque: blood-lumen/fluid domain; translucent: CATIA-generated wall domain. Renderings, not patient photographs.}
\label{fig1}
\end{figure}

\begin{figure*}[htbp]
\centering
\includegraphics[width=0.8\linewidth]{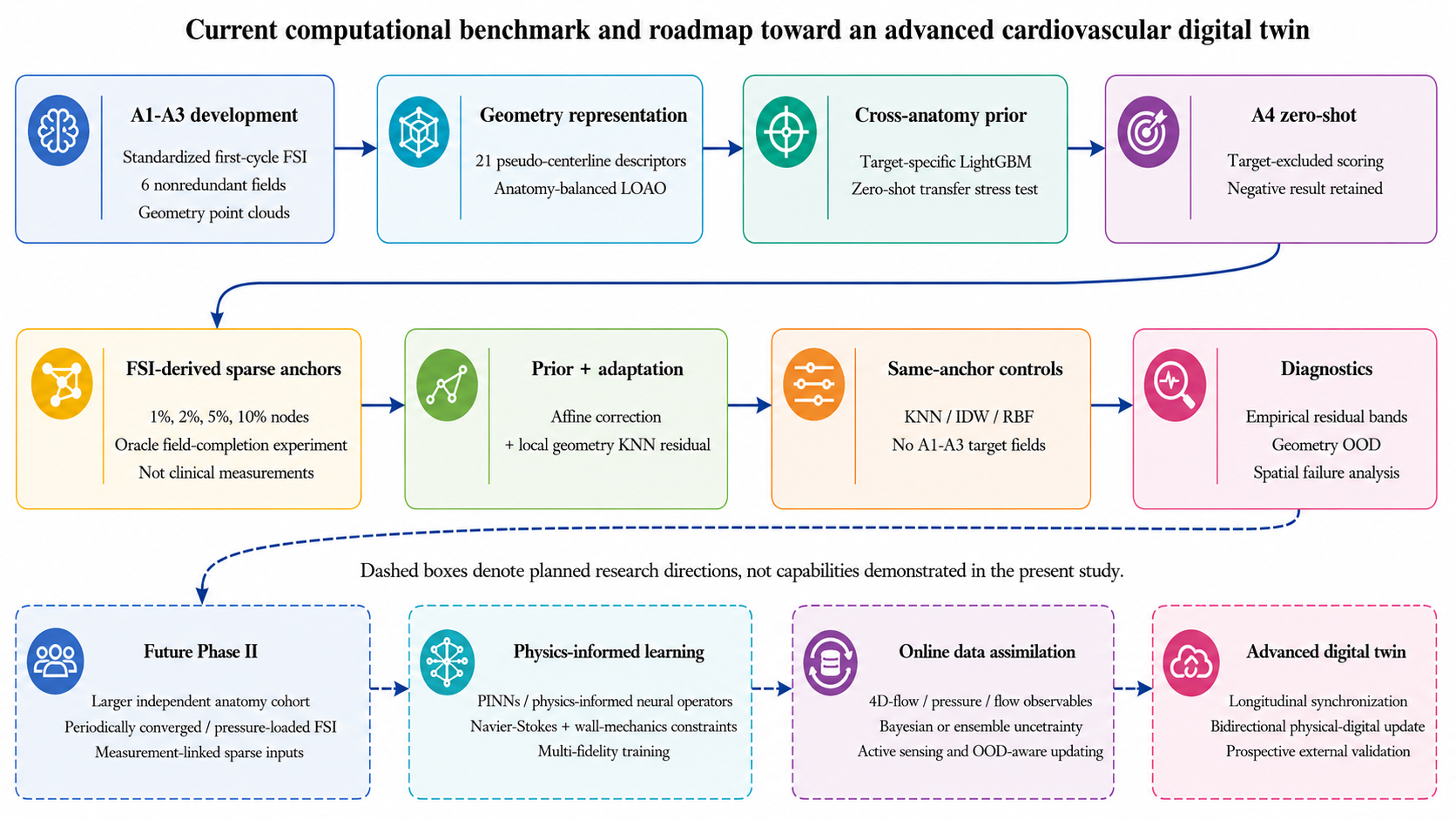}
\caption{Evaluation workflow and research roadmap. Solid boxes: capabilities evaluated here. Dashed boxes: planned extensions, not current results.}
\label{fig2}
\end{figure*}

Geometry preparation followed a deterministic CAD workflow: VMR VTP models were imported into ParaView, exported as STL surfaces, then imported into CATIA to construct separate blood-lumen and wall solids. A nominal 1.5-mm wall thickness was imposed during reconstruction, a modeling assumption, not an image-measured quantity. Separate fluid/wall STEP solids were exported from CATIA into COMSOL for FSI. Fig.~\ref{fig1} shows representative A1--A3 development domains.

\subsection{Anatomy-Specific FSI Under Standardized Assumptions}
Each anatomy comprised volumetric fluid and wall domains, solved in COMSOL Multiphysics 6.4 with two-way FSI and an arbitrary Lagrangian--Eulerian moving fluid mesh. Geometry was individualized, while inflow, outlet pressure, and material parameters were standardized: this is anatomy-specific FSI under standardized conditions, not physiologically patient-specific FSI.

The incompressible fluid equations were the continuity condition $\nabla \cdot \mathbf{u} = 0$ and
\begin{equation}
\rho_f \left[\frac{\partial \mathbf{u}}{\partial t} + (\mathbf{u} - \mathbf{w}) \cdot \nabla \mathbf{u}\right] = -\nabla p + \nabla \cdot \boldsymbol{\tau}, \label{eq2}
\end{equation}
with $\rho_f = 1060~\text{kg/m}^3$. Carreau viscosity was
\begin{equation}
\mu(\dot{\gamma}) = \mu_\infty + (\mu_0 - \mu_\infty)\left[1 + (\lambda \dot{\gamma})^2\right]^{(n-1)/2}, \label{eq3}
\end{equation}
where $\mu_0 = 0.056~\text{Pa\,s}$, $\mu_\infty = 0.0035~\text{Pa\,s}$, $\lambda = 3.313~\text{s}$, and $n = 0.3568$ \cite{b7}. The wall used a two-parameter Mooney--Rivlin law $W_{\text{iso}} = C_{10}(I_1 - 3) + C_{01}(I_2 - 3)$ with $C_{10} = 0.174$~MPa, $C_{01} = 1.88$~MPa, solid density 1200~kg/m$^3$, bulk modulus 6.667~MPa, benchmark settings, not calibrated to tissue tests. Hyperelastic wall modeling agrees with arterial-mechanics literature, though absolute aneurysm wall stress stays sensitive to constitutive/geometric assumptions \cite{b20,b21}. No prestress, zero-pressure-geometry correction, or external pressure load was applied; wall end regions were fixed to remove rigid-body motion, fluid--wall interface stayed coupled.

All cases used the same one-second inlet waveform and a 0-Pa-gauge outlet; exact velocity knots, spline interpolation, and moving-mesh parameters are in the Supplement. Initial pressure was 0~Pa with zero initial velocity/displacement, so the 0--1~s interval includes startup from an unloaded/rest state. Backward differentiation formula (BDF) order 1--2, relative tolerance $10^{-3}$, and maximum internal step 0.005~s were common; 21 fields were exported at 0.05-s spacing, with two boundary-layer elements at the wall. Final tetrahedral counts were 278,889, 213,130, 415,816, and 128,656, minimum quality 0.145, 0.077, 0.121, and 0.144, the lower A2 value is a numerical limitation, not mesh adequacy. The A2 solution stored the complete state through exactly $t = 1.0$~s before a later step failed.

Formal mesh-refinement and time-step convergence series could not be regenerated within the archived campaign, so we do not label the fields mesh- or time-step-independent. All anatomies shared the same meshing strategy, boundary-layer elements, BDF order, tolerance, and step size, so central claims are comparative, not claims of converged stress. An output-sampling check (0.05-s grid vs. every-second-state recomputation) gave median TAWSS differences of 2.2--3.6\% across anatomies, OSI 15.6--28.1\%, and 95th-percentile peak-WSS 17.9--30.1\%, supporting the finer grid without replacing formal convergence testing.

\subsection{Vector Wall Shear Stress and Mechanical Targets}
From apparent viscosity, velocity gradients, and unit normal $\mathbf{n}$, the viscous tensor and tangential traction were reconstructed as
\begin{equation}
\boldsymbol{\tau} = \mu_{\text{app}}(\nabla \mathbf{u} + \nabla \mathbf{u}^T), \quad \boldsymbol{\tau}_w = \boldsymbol{\tau}\mathbf{n} - [(\boldsymbol{\tau}\mathbf{n}) \cdot \mathbf{n}]\mathbf{n}. \label{eq5}
\end{equation}
Trapezoidal quadrature over the 21 exported states gave time-averaged wall shear stress $\text{TAWSS} = T^{-1}\int_0^T |\boldsymbol{\tau}_w|\,dt$, oscillatory shear index
\begin{equation}
\text{OSI} = \frac{1}{2}\left(1 - \frac{\left|\int_0^T \boldsymbol{\tau}_w\,dt\right|}{\int_0^T |\boldsymbol{\tau}_w|\,dt}\right), \label{eq7}
\end{equation}
and relative residence time $\text{RRT} = 1/[(1-2\,\text{OSI})\,\text{TAWSS}]$. Peak WSS was the maximum magnitude among the 21 exported states (a saved-state maximum, not over internal solver steps); structural targets were the maximum among the same states and the trapezoidal temporal mean of von Mises stress. All target tables were audited for missing/non-finite values; none were present. RRT was not capped: its largest value reached $1.79 \times 10^4~\text{Pa}^{-1}$ in anatomy 4, motivating the pre-specified $\log(1+y)$ transform and rank-based reporting.

A seventh candidate target, stress amplitude, was dropped: since simulations start from an effectively unloaded state, minimum von Mises stress was near zero everywhere, so amplitude correlated with cycle maximum at $r = 1.000000$ and equaled $0.5\,\sigma_{\max}$ within relative deviation below $10^{-7}$ across anatomies, and treating both as independent would double-count one field.

Reconstructed WSS was checked against native COMSOL FSI tangential traction on two anatomies via slope/intercept, bias, MAE, IQR-normalized RMSE, and percentile errors. For anatomy 3, TAWSS slope was 1.00015, intercept $5.19\times10^{-4}$~Pa, MAE 0.00102~Pa, IQR-normalized RMSE 0.0119, median/95th-percentile relative errors 0.065\%/0.579\%. For anatomy 4, slope was 1.00135, intercept $-8.33\times10^{-4}$~Pa, MAE 0.00126~Pa, IQR-normalized RMSE 0.0094, median/95th-percentile errors 0.033\%/0.351\%. OSI MAE was $1.15\times10^{-4}$/$1.62\times10^{-4}$; RRT was small-error for most nodes but unstable in raw scale for anatomy 4, where near-singular values dominate squared error, validating post-processing consistency, not FSI physiological correctness.

\subsection{Geometry-Only Multi-Anatomy Prior}
Each node was represented by 21 geometry descriptors from the inner-wall point cloud. Rather than an independently segmented anatomical centerline, a deterministic point-cloud pseudo-centerline was built via principal-axis projection, 24 axial quantile bins, bin-centroid smoothing, and nearest-centerline assignment. Predictors included normalized pseudo-centerline position, nearest-end distance, normalized radial position, local-radius ratios, radius gradient, curvature/curvature-side coordinates, tangent alignment, length, radius-to-length and expansion/end-radius ratios, tortuosity, curvature--length summaries, and two bounding-box aspect ratios (full definitions in the Supplement).

Transforms were fixed as log1p for TAWSS, RRT, and peak WSS; identity for OSI; natural log for wall stresses. Three target-specific LightGBM complexity levels \cite{b23} were compared via three-fold LOAO, with anatomy-balanced node weights preventing the densest mesh from dominating fitting. Candidate selection minimized mean LOAO normalized MAE, defined per held-out anatomy as $\text{MAE}/(Q_{0.75} - Q_{0.25} + 10^{-12})$, Spearman correlation secondary. The final-fit script then overrode the candidate-specific tree count with 450 trees; this pre-A4 detail and its sensitivity audit are in the Supplement.

\subsection{Sparse Field Completion and Same-Anchor Controls}
Sparse calibration is an oracle simulation experiment: anchor labels are FSI-derived target values, not clinical measurements, and wall-stress anchors have no direct measurement analogue. It asks how well the remaining field can be completed given a small known subset; it does not demonstrate end-to-end clinical updating or a speedup when the labels themselves require full FSI.

Anchor sets of 1, 2, 5, or 10\% were chosen with MiniBatchKMeans over eight standardized local-geometry features. Since clustering uses geometry from all nodes but no non-anchor target values, the experiment is transductive. The prior-based method first fit a ridge affine correction ($\alpha = 10^{-3}$) from transformed prior predictions to transformed anchor labels, then a distance-weighted $k=10$ $k$-nearest-neighbor (KNN) model to remaining anchor residuals; all performance calculations excluded anchor nodes.

To answer RQ3, three controls used the same target anchors but no A1--A3 target fields, with hyperparameters selected only on A1--A3 5\% LOAO development over a pre-specified grid: (i) distance-weighted KNN, $k=5$, in the eight local-geometry features; (ii) inverse-distance weighting (IDW) from the eight nearest anchors in standardized Cartesian coordinates, weights $d^{-3}$; (iii) thin-plate-spline radial basis function (RBF) interpolation, smoothing $10^{-3}$, at most 80 local neighbors. These were fixed for the secondary A4 comparisons.

\subsection{Metrics, Dependence Sensitivities, and Empirical Bands}
Primary metrics were $R^2$, MAE, RMSE, Spearman $\rho$, and top-decile overlap. For $N$ non-anchor evaluation nodes, $k = \text{round}(0.1N)$ and $\text{Top10} = |H_{\text{ref}} \cap H_{\text{pred}}|/k$, each set containing exactly the $k$ largest values; ties at the cutoff were resolved by fixed array order. RRT was additionally evaluated in $\log(1+\text{RRT})$ space, since its raw distribution is singular as $(1-2\,\text{OSI})\,\text{TAWSS} \to 0$. Any ``six-target mean'' is an unweighted macro-average across heterogeneous tasks; no inferential $p$-values treat wall nodes as independent subjects.

Primary metrics are node-weighted, since finite-element (FE) vertex-area weights were not preserved in the archived CSV exports. To test mesh-density sensitivity, we added a local-spacing weighting analysis (weight proportional to mean squared distance to eight nearest surface nodes), plus five equal-count centerline blocks on the fourth anatomy.

Because MiniBatchKMeans is stochastic, anchor-selection robustness at the 5\% setting was checked against 10 alternative KMeans seeds and 10 uniformly random anchor sets, using the development-selected IDW interpolator across the six primary tasks.

For secondary uncertainty analysis, the sparse-calibration procedure was replayed under A1--A3 LOAO: for each target, the largest of the three held-out-anatomy 90th-percentile absolute residuals in transformed space defined a development-calibrated empirical residual band, not a distribution-free 90\% predictive interval. Local geometry shift was quantified via pair-calibrated 10-nearest-neighbor distances in the same eight features. Prior-based OSI predictions stayed within $[0, 0.5]$ without clipping; the unconstrained RBF baseline produced 21 out-of-range OSI predictions among 3,852 non-anchor A4 nodes at 5\% (0.55\%; minimum $-0.0122$, maximum 0.5180). A clipped $[0, 0.5]$ sensitivity changes OSI $R^2$ only from 0.9166 to 0.9171.

\begin{table}[htbp]
\caption{Computational Anatomy and FSI Summary. Pseudo-Centerline Length/Tortuosity Are Geometry-Descriptor Values, Not Clinical Measurements.}
\label{tab1}
\centering

\scriptsize
\setlength{\tabcolsep}{3pt}
\renewcommand{\arraystretch}{1.05}

\resizebox{\columnwidth}{!}{%
\begin{tabular}{|l|c|c|c|c|}
\hline
\textbf{Metric} & \textbf{A1} & \textbf{A2} & \textbf{A3} & \textbf{A4} \\
\hline
Wall nodes & 8191 & 6800 & 3203 & 4055 \\
\hline
Tetrahedra & 278889 & 213130 & 415816 & 128656 \\
\hline
Min tet. quality & .145 & .077 & .121 & .144 \\
\hline
Pseudo-centerline length (mm) & 128.09 & 96.21 & 93.69 & 85.40 \\
\hline
Tortuosity (geom.) & 1.226 & 1.256 & 1.138 & 1.346 \\
\hline
Mean TAWSS (Pa) & .964 & 1.218 & .699 & 1.144 \\
\hline
P95 TAWSS (Pa) & 2.380 & 4.559 & 1.094 & 2.878 \\
\hline
Global saved-state peak WSS (Pa) & 74.73 & 134.27 & 33.31 & 101.79 \\
\hline
P95 saved-state peak WSS (Pa) & 14.14 & 27.41 & 5.04 & 15.55 \\
\hline
Mean saved-state max VM (kPa) & 3.916 & 2.472 & 3.140 & 2.518 \\
\hline
P95 saved-state max VM (kPa) & 10.55 & 4.76 & 5.92 & 4.87 \\
\hline
\end{tabular}%
}
\end{table}

\section{Results}

\subsection{Computational-Reference Audits}
Table~\ref{tab1} summarizes the four anatomies; robust percentiles are reported alongside global maxima since the latter are mesh-sensitive. Hemodynamic and mechanical rankings differed: anatomy 2 had the highest global saved-state peak WSS (134.27~Pa) and 95th-percentile TAWSS (4.56~Pa), while anatomy 1 had the largest 95th-percentile saved-state maximum von Mises stress (10.55~kPa), heterogeneity supporting separate interpretation of flow and mechanical fields over a single fixed-weight ``risk'' score.

The WSS agreement audit showed near-unity slopes and millipascal-scale TAWSS bias/MAE on anatomies 3 and 4, supporting the reconstruction implementation. The stress-amplitude audit resolved the duplicated-looking metrics: amplitude was effectively half of cycle maximum, since minimum stress was near zero under first-cycle unloaded initialization, so it was dropped from all primary averages and tables.

\subsection{Research Question 1: Zero-Shot Cross-Anatomy Transfer Is Weak}
On anatomy 4, the geometry-only prior gave $R^2 = -0.159$ (TAWSS), $-0.374$ (OSI), $\approx 0$ (RRT), $-0.027$ (peak WSS), $-0.095$ (maximum von Mises stress, VM), $-0.056$ (mean VM): uniformly poor. A source-code audit found the final refit used 450 trees for every target; re-fitting with each candidate's native tree count moved macro zero-shot $R^2$ from $-0.119$ to $-0.089$ and macro $\rho$ from 0.115 to 0.133, both still poor, with 5\% adapted performance essentially unchanged, the negative result does not hinge on the tree-count override, directly answering RQ1: with only three development anatomies, this hand-crafted geometry representation does not support reliable zero-shot transfer.

\subsection{Research Question 2: Sparse Labels Improve Reconstruction of Non-Anchor Nodes}
\begin{figure}[htbp]
\centering
\includegraphics[width=\linewidth]{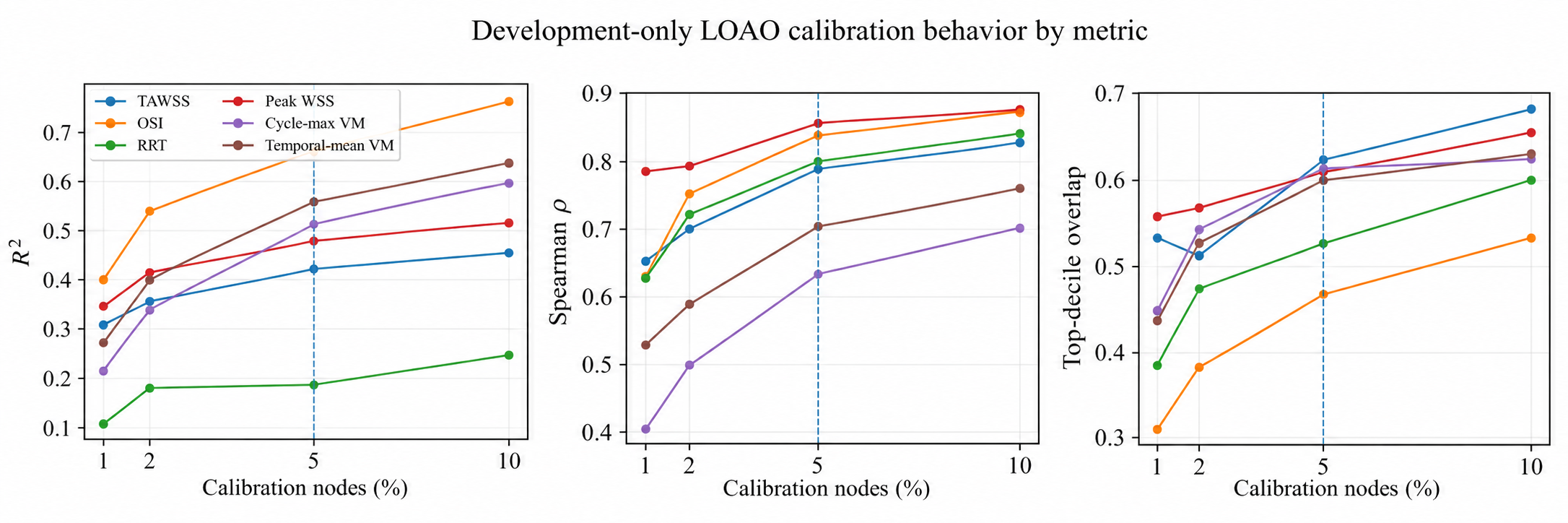}
\caption{Development-only LOAO calibration behavior for the six primary targets ($R^2$, Spearman $\rho$, top-decile overlap). Vertical line: preselected 5\% setting.}
\label{fig3}
\end{figure}

Development LOAO curves improved monotonically with additional within-anatomy labels (Fig.~\ref{fig3}), though target-specific behavior was heterogeneous. The 5\% setting was retained as primary calibration, showing a clear development-fold gain without doubling the burden to 10\%. On anatomy 4 (203 anchors, 3,852 evaluation nodes), prior-plus-adaptation reached OSI $R^2 = 0.603$, $\rho = 0.807$, 0.668 top-decile overlap; peak WSS $R^2 = 0.302$, $\rho = 0.732$; TAWSS $R^2 = 0.234$, $\rho = 0.671$; saved-state maximum VM $R^2 = 0.317$; temporal-mean VM $R^2 = 0.410$. RRT raw-scale $R^2$ stayed near zero, but $\rho = 0.711$, top-decile overlap 0.579. Table~\ref{tab2} gives the full error profile (mechanical MAE/RMSE in kPa); the empirical-band column is descriptive, not a formal guarantee.

\begin{table}[htbp]
\caption{Prior-Plus-Adaptation Performance on Anatomy 4, 5\% Setting (Non-Anchor Nodes Only).}
\begin{center}
\resizebox{\linewidth}{!}{%
\begin{tabular}{|l|c|c|c|c|c|c|}
\hline
\textbf{Target} & \textbf{$R^2$} & \textbf{MAE} & \textbf{RMSE} & \textbf{$\rho$} & \textbf{Top10} & \textbf{Emp. band} \\
\hline
TAWSS (Pa) & .234 & .408 & 1.179 & .671 & .587 & .884 \\
\hline
OSI & .603 & .0438 & .0645 & .807 & .668 & .914 \\
\hline
RRT (Pa$^{-1}$) & .000 & 5.959 & 287.697 & .711 & .579 & .900 \\
\hline
Saved-state peak WSS (Pa) & .302 & 2.035 & 5.835 & .732 & .605 & .894 \\
\hline
Saved-state max VM (kPa) & .317 & .729 & 1.029 & .569 & .506 & .903 \\
\hline
Temporal-mean VM (kPa) & .410 & .170 & .239 & .673 & .506 & .901 \\
\hline
\end{tabular}}
\label{tab2}
\end{center}
\end{table}

\subsection{Research Question 3: Same-Anchor Interpolation Matches or Outperforms Prior-Based Adaptation}
The critical result: sparse-field success cannot be attributed to the multi-anatomy prior. With direct-baseline hyperparameters selected exclusively from A1--A3, 5\% A4 calibration gave six-task macro $R^2 = 0.431$ (IDW), 0.425 (RBF), versus 0.311 (prior-plus-adaptation); macro Spearman 0.800/0.848 versus 0.694. RBF was especially strong for OSI ($R^2 = 0.917$), maximum VM ($R^2 = 0.714$), and mean VM ($R^2 = 0.778$), while the prior stayed more competitive for TAWSS/peak-WSS (Table~\ref{tab3}); no single rule dominated, but the cross-anatomy prior gave no consistent incremental value.

The same conclusion held without A4 in baseline selection. In A1--A3 5\% LOAO development, six-task macro $R^2$ was 0.530 (selected IDW), 0.572 (selected RBF), versus 0.469 (prior-plus-adaptation); macro $\rho$ was 0.853, 0.881, 0.772. Selected IDW/RBF exceeded prior-plus-adaptation macro $R^2$ in each held-out anatomy (A1: 0.555/0.576 vs. 0.553; A2: 0.587/0.665 vs. 0.529; A3: 0.446/0.475 vs. 0.327), so the gap is not one anatomy's artifact, and held across A4 calibration fractions 1--10\%. Sparse labels are informative, but direct interpolation must be benchmarked before crediting performance to cross-anatomy transfer.

Spatial maps (Fig.~\ref{fig5}) show the same result: RBF reconstructs OSI and maximum VM particularly well but is less stable for peak WSS and, being unconstrained, yields a few out-of-range OSI values; IDW, bounded by its anchor-value convex combination, is more uniform across targets, arguing for target-specific baseline selection over one universal algorithm.

\begin{table}[htbp]
\caption{Anatomy-4 Reconstruction at 5\% Calibration: Prior-Plus-Adaptation vs. Direct Baselines. RRT $R^2$ Shown as 0.000 to Avoid Negative-Zero Formatting.}
\begin{center}
\resizebox{\linewidth}{!}{%
\begin{tabular}{|l|c|c|c|c|c|c|c|c|}
\hline
\multirow{2}{*}{\textbf{Target}} & \multicolumn{2}{c|}{\textbf{Prior+adapt}} & \multicolumn{2}{c|}{\textbf{KNN anchors}} & \multicolumn{2}{c|}{\textbf{IDW anchors}} & \multicolumn{2}{c|}{\textbf{RBF anchors}} \\
\cline{2-9}
 & $R^2$ & $\rho$ & $R^2$ & $\rho$ & $R^2$ & $\rho$ & $R^2$ & $\rho$ \\
\hline
TAWSS & .234 & .671 & .228 & .679 & .185 & .747 & .133 & .785 \\
\hline
OSI & .603 & .807 & .664 & .835 & .829 & .899 & .917 & .950 \\
\hline
RRT & .000 & .711 & .000 & .731 & .000 & .783 & .000 & .835 \\
\hline
Saved-state peak WSS & .302 & .732 & .299 & .747 & .277 & .784 & .013 & .815 \\
\hline
Saved-state max VM & .317 & .569 & .359 & .573 & .617 & .754 & .714 & .823 \\
\hline
Temporal-mean VM & .410 & .673 & .439 & .677 & .676 & .832 & .778 & .878 \\
\hline
\end{tabular}}
\label{tab3}
\end{center}
\end{table}

\subsection{Research Question 4: Empirical Residual Bands and Geometry Shift Are Diagnostic, Not Guarantees}
Development-calibrated empirical residual bands covered 88.4\% (TAWSS), 91.4\% (OSI), 90.0\% (RRT), 89.4\% (peak WSS), 90.3\% (maximum VM), 90.1\% (mean VM) of non-anchor anatomy-4 values, but band widths were large: median 2.05 development IQRs (TAWSS), 1.77 (peak WSS), 1.53/1.63 (max/mean VM). The same fixed worst-fold radii gave 89.98--96.65\% coverage across A1--A3 LOAO folds, descriptive, not general predictive calibration.

\begin{figure}[htbp]
\centering
\includegraphics[width=\linewidth]{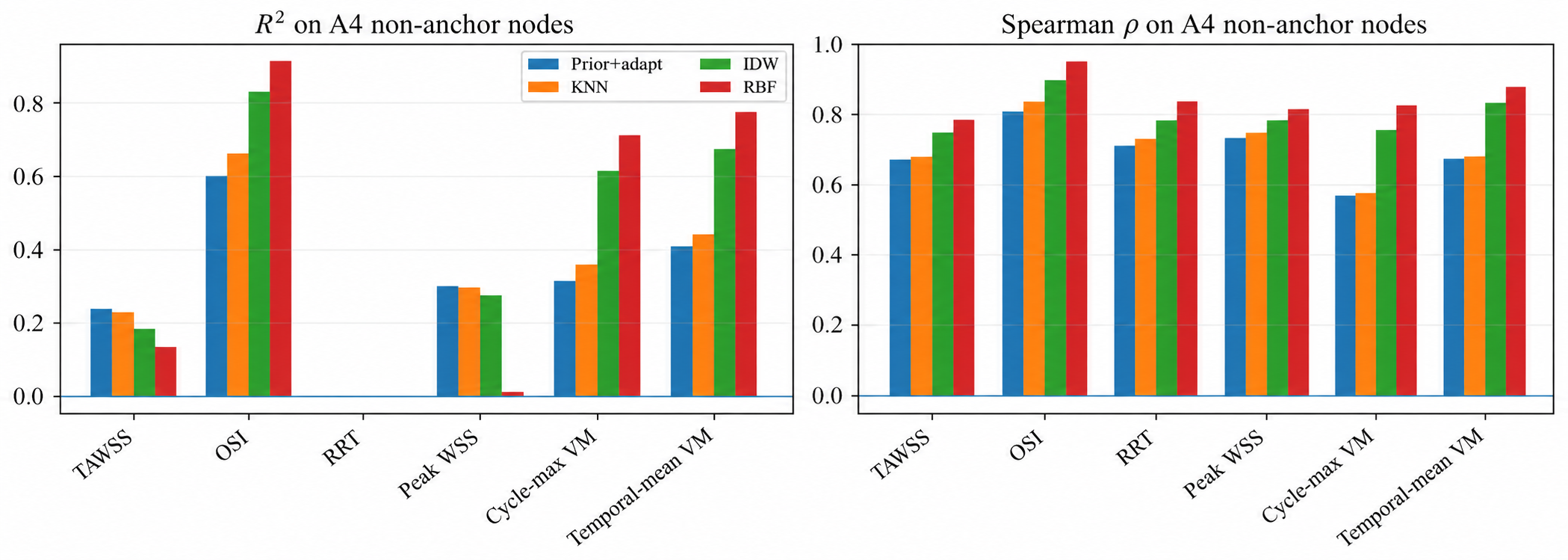}
\caption{Same-anchor controls on anatomy 4 at 5\% calibration. Direct IDW/RBF interpolation frequently equal or exceed prior-plus-adaptation.}
\label{fig4}
\end{figure}

\begin{figure}[htbp]
\centering
\includegraphics[width=\linewidth]{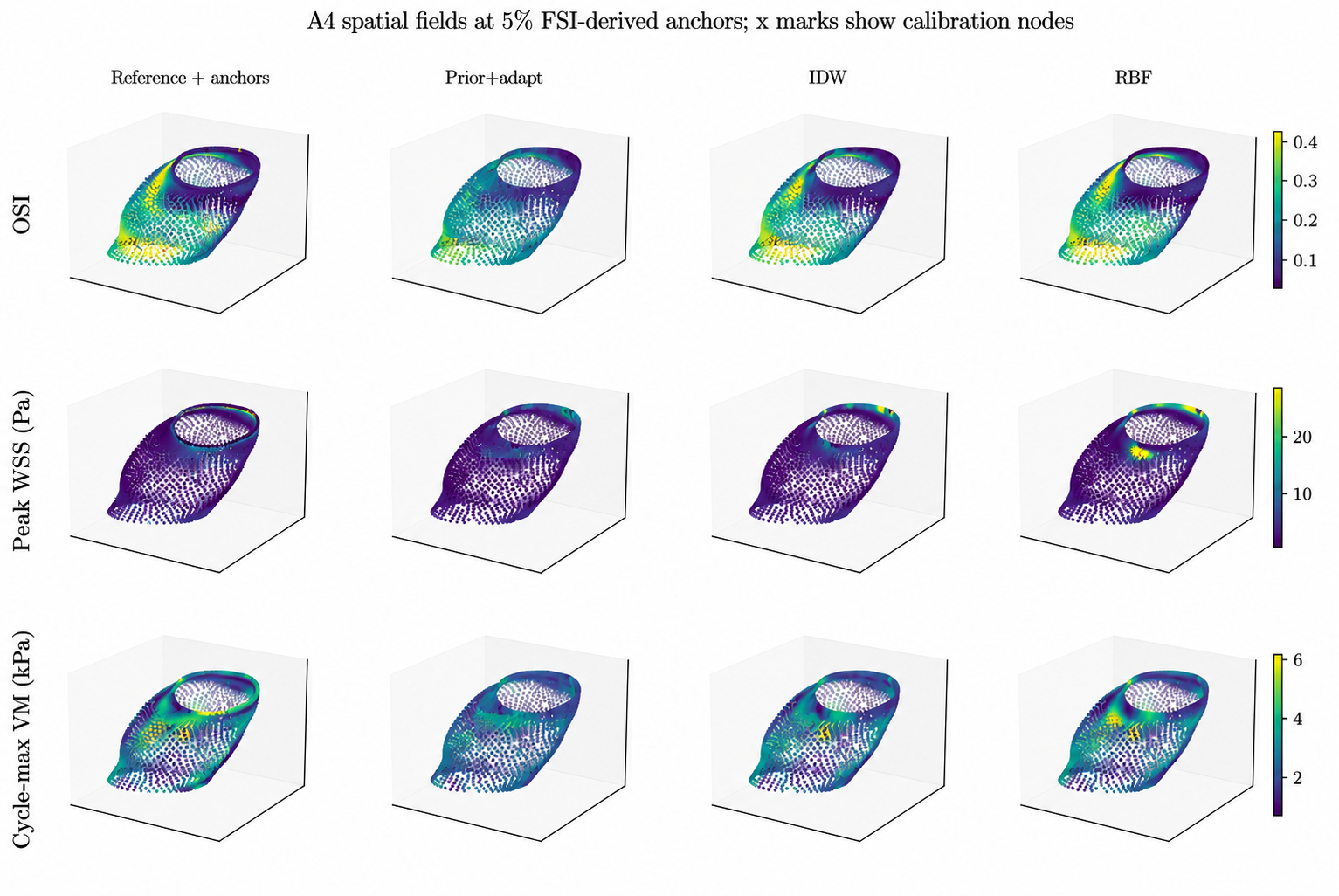}
\caption{Anatomy-4 non-anchor spatial fields at 5\% calibration. Rows: OSI, saved-state peak WSS, saved-state maximum VM stress. Columns: FSI reference, prior-plus-adaptation, IDW, RBF; matched display limits per row.}
\label{fig5}
\end{figure}

\begin{figure*}[htbp]
\centering
\includegraphics[width=0.9\linewidth]{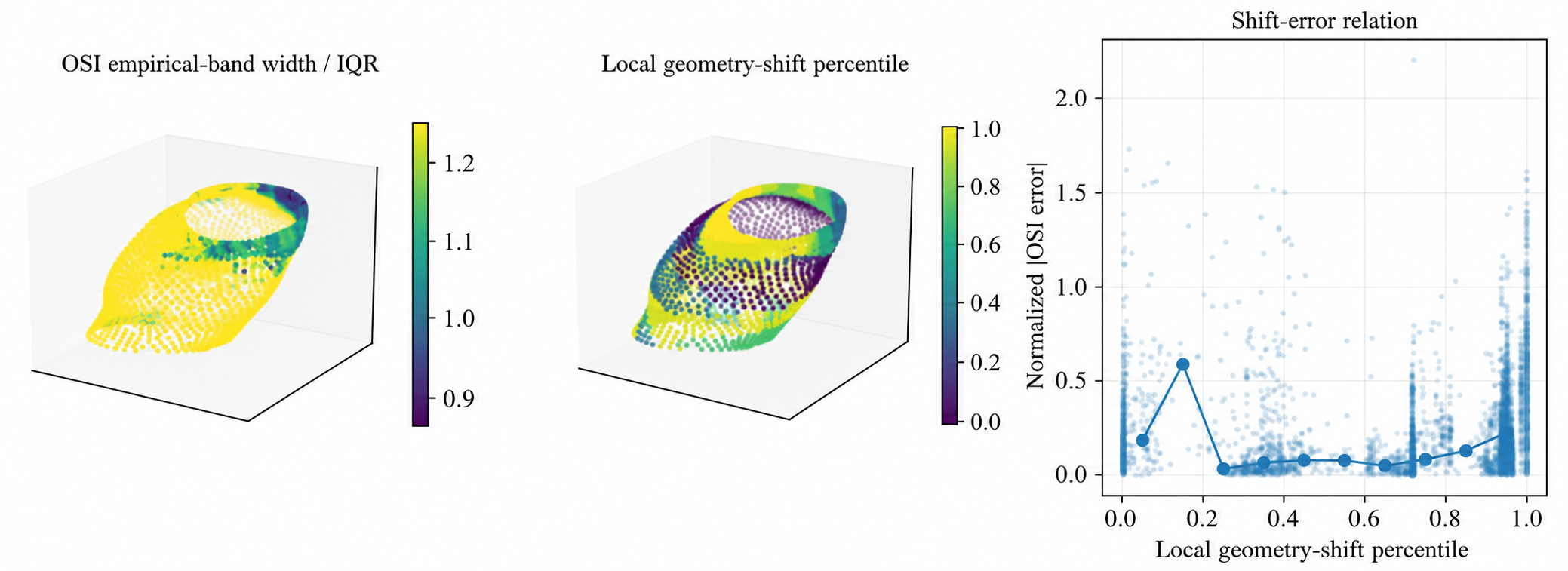}
\caption{Secondary diagnostics on non-anchor anatomy-4 nodes: empirical residual-band width for OSI (normalized by development IQR) and local geometry-shift percentile. Not clinically calibrated.}
\label{fig6}
\end{figure*}

The median local geometry-shift percentile on anatomy 4 was 0.808 (20.5\% of nodes exceeded the 95th percentile, 11.0\% the 99th). Association with normalized absolute error was target-dependent ($\rho = 0.292$ OSI, 0.224 RRT, 0.151 TAWSS) and negligible for maximum wall stress: the out-of-distribution (OOD) score is a geometry-support marker, not an error oracle (Fig.~\ref{fig6}).

\subsection{Dependence and Mesh-Density Sensitivities}
Node weighting did not create the central baseline result: recomputing the local-surface-density sensitivity (a proxy, not exact FE quadrature) with A1--A3-selected settings preserved the qualitative ordering for OSI and both mechanical fields, though RBF stayed unstable for absolute TAWSS/peak-WSS $R^2$. Five longitudinal pseudo-centerline blocks showed substantial block-to-block variability (prior-plus-adaptation TAWSS block $R^2$ ranged 0.086--0.633, IDW $-0.032$--0.762, RBF $-1.979$--0.895), reinforcing that thousands of mesh nodes are not independent replicates.

\subsection{Anchor-Selection Robustness}
The central IDW result was not one favorable KMeans seed: across 10 alternative 5\% seeds, six-task IDW macro $R^2$ on A1--A3 LOAO averaged $0.546\pm0.093$, macro $\rho = 0.845\pm0.060$, versus $0.500\pm0.135$/$0.823\pm0.067$ for random same-size anchor sets; on A4, KMeans gave $0.422\pm0.025$/$0.792\pm0.012$ versus $0.389\pm0.042$/$0.762\pm0.016$ for random anchors, representative selection improves stability modestly, but the same-anchor conclusion is not a single-seed artifact. A nearest-anchor-distance audit showed the expected transductive pattern: macro median absolute error normalized by target IQR was 0.157, 0.141, 0.148, 0.200 nearest-to-farthest (IDW), 0.120, 0.093, 0.106, 0.157 (RBF), not strictly monotone, so distance is a support diagnostic, not an error model.

\subsection{Computational Cost Context}
Archived COMSOL logs give an order-of-magnitude comparison: final-physics logs for A2--A4 recorded approximately 18,968~s (5.27~h), 5,251~s (1.46~h), and 5,499~s (1.53~h); A1's runtime was not recoverable. A post-hoc Python timing audit (five CPU cores) required median 0.0046~s for A4 geometry-feature construction, 0.074~s for LightGBM inference, 0.088~s for 5\% KMeans anchor selection, 0.052~s for prior adaptation, 0.0054~s for IDW completion, and 1.68~s for local RBF completion over 4,055 nodes, negligible next to the archived FSI solves, but no end-to-end clinical speedup is established, since anchors are oracle FSI values and CAD/meshing, FSI generation, and measurement acquisition sit outside this audit.

\section{Discussion}

\subsection{What the Baseline Changes}
The most consequential revision is RQ3. The original multi-anatomy framing could have read improved sparse-calibration performance as evidence that an anatomy-trained prior became useful after personalization; same-anchor controls show this is not supported, and the gap widens when interpolation hyperparameters are selected only on A1--A3 and fixed. The defensible conclusion is narrower but more informative: sparse within-anatomy field completion is feasible for several smooth FSI-derived fields, but the four-anatomy prior adds little demonstrable value beyond those labels.

This matters for the surrogate literature: large geometry cohorts can learn cross-shape field relationships when anatomical coverage is broad \cite{b9}, and reduced-order/operator approaches work well when training support matches the test distribution \cite{b10,b11,b14}. Our negative zero-shot result and strong same-anchor controls show the opposite regime, very few anatomies and a smooth spatial field, where complexity should not be credited unless it beats a direct within-anatomy baseline.

\subsection{Meaning of the Sparse Anchors}
The 5\% anchors are not a practical clinical operating point: they are oracle labels from the full FSI reference, used to study field completion. Within the digital-twin-oriented design, this is an upper-bound experiment for the update layer, asking whether sparse state information could condition a full-field surrogate if it became available from measurements. Sparse TAWSS/OSI may eventually come from imaging-derived flow measurements, but wall stress is not directly measured here, and if anchor values require a full FSI run, the experiment saves no computation, a proof-of-concept sparse-update module, not a demonstrated clinical synchronization mechanism.

\subsection{Numerical and Physiological Scope}
The surrogate cannot be more physiologically valid than its reference fields. The study deliberately uses one matched startup cycle, common inflow, 0-Pa-gauge outlet pressure, common material constants, no prestress, and no external pressure so geometry and surrogate behavior compare under a controlled benchmark; this reduces between-case confounding but does not make absolute wall stresses physiological. Wall-mechanical quantities are thus computational response fields, not rupture stresses, and TAWSS/OSI/RRT are first-cycle, not periodically converged, metrics. Coarsening the saved VM series from 0.05-s to 0.10-s spacing shifted median saved-state maximum VM by 16.5--33.4\% (95th-percentile 49.2--55.7\%), while temporal-mean VM was more stable (7.5--12.8\%). Formal mesh refinement, convergence testing, multi-cycle periodicity, and physiologic pressure/prestress sensitivity remain necessary before clinical interpretation \cite{b8}; their absence does not invalidate the narrower algorithm-comparison question addressed here.

\subsection{Why the Current Cross-Anatomy Prior Underperforms}
No single cause explains weak zero-shot transfer, but several mechanisms narrow the picture. Only three anatomies inform development, so the geometry-to-field relation is severely under-sampled. The 21 descriptors impose correspondence via a principal component analysis (PCA)/binning pseudo-centerline rather than true anatomical correspondence, so tortuous or asymmetric regions can land in different feature neighborhoods across anatomies. LightGBM predicts from pointwise descriptors with no explicit surface connectivity or FSI interface physics, and physiology is fixed across anatomies, so the model learns only geometry-conditioned variation. Finally, the strong IDW/RBF controls show several targets are locally smooth once sparse labels exist, giving local interpolation an inherent edge over a tiny learned prior, motivating models that encode geometry and physics directly.

\subsection{Roadmap Toward an Advanced Digital Twin}
The architecture is deliberately digital-twin-oriented, but its demonstrated scope remains the computational surrogate/update layer, not a complete clinical twin. We view this as Phase I of a continuing program: it establishes an FSI reference workflow, exposes the limits of geometry-only cross-anatomy transfer, quantifies sparse within-anatomy information, and provides falsifiable same-anchor controls. Modern digital twins require a dynamic link to a physical counterpart and real-world data assimilation \cite{b16}; recent cardiovascular examples infer individualized physiology from wearable data \cite{b15,b17}.

Planned next stages will first expand and numerically validate the FSI cohort, then replace pointwise regression with physics-informed neural/operator or mesh/graph architectures \cite{b24} encoding Navier--Stokes residuals, wall constitutive behavior, interface continuity, and multi-fidelity constraints. A later translational stage replaces oracle FSI anchors with measurable signals (4D-flow MRI, pressure/flow waveforms, serial image-derived morphology), with Bayesian/ensemble uncertainty, OOD-aware abstention, active sensing, longitudinal updating, and prospective external validation. The long-term goal is a synchronized cardiovascular digital twin with measurement-conditioned, physics-informed inference, explicit future directions, not claims strengthening current results.

\section{Limitations}
Effective sample size is four anatomies, not 22,249 nodes: only three inform prior development, and A4 sparse-calibration results are post-zero-shot case studies, not pristine external validation. The FSI reference is first-cycle, standardized in boundary/material inputs, without prestress or convergence testing. Sparse anchors are oracle simulation values, not clinical observables, and the completion experiment is transductive. Direct interpolation outperformed prior-based adaptation on several targets, so no transfer-learning advantage is established; unconstrained RBF can violate the OSI range, exact FE area weights were unavailable, and empirical bands/repeated-seed checks carry no formal statistical guarantee. RRT is heavy-tailed with unstable raw-scale $R^2$; WSS agreement validates only internal post-processing, restricting the study to comparative computational-surrogate conclusions, no rupture-risk or clinical claim.

\section{Data, Code, and Declarations}
The source models were downloaded from the Vascular Model Repository\footnote{https://www.vascularmodel.com/dataset.html}, a public repository of de-identified cardiovascular imaging and model data \cite{b3}. This is a secondary computational analysis of publicly released model files; no participants were recruited, no human subjects contacted, no direct identifiers accessed. The geometry-processing path (VTP to ParaView/STL, to CATIA lumen/wall solids, to STEP, to COMSOL) is documented in this paper and Supplement. The VMR license permits research/development use under its stated copyright and acknowledgement conditions; source data should be obtained from VMR under those terms. The reproducibility package contains the analysis code, freeze manifest/hashes, processed feature/target tables, LOAO/baseline outputs, and sensitivity analyses. The exact VMR identifier-to-A1--A4 mapping is incomplete in derived filenames. The authors declare no conflict of interest.

\section{Conclusion}
This four-anatomy study separates three questions often conflated in scientific ML for FSI: node-level interpolation, cross-anatomy transfer, and sparse within-anatomy field completion. The geometry-only multi-anatomy prior failed in zero-shot transfer, and while sparse FSI-derived labels improved reconstruction, direct same-anchor IDW/RBF interpolation matched or exceeded the prior-based method on average and in development LOAO folds alike, the improvement is attributable to within-anatomy information, not demonstrated cross-anatomy transfer. The contribution is thus a digital-twin-oriented surrogate evaluation, not a completed clinical twin: simple baselines show when a sophisticated personalization narrative is unnecessary, while geometry-conditioned surrogacy, sparse state updating, and explicit failure controls mark this as the first stage of a continuing program. Next stages require more independent anatomies, exact surface-area evaluation, converged and pressure-loaded FSI, measurement-linked inputs, and physics-informed neural/operator models embedding governing fluid/solid equations; the long-term goal is a synchronized cardiovascular digital twin with longitudinal data assimilation and prospective external validation, capabilities this study stops short of claiming.


\bibliographystyle{IEEEtran}
\bibliography{references}

@article{b1,
  author  = {Vorp, David A.},
  title   = {Biomechanics of abdominal aortic aneurysm},
  journal = {Journal of Biomechanics},
  year    = {2007},
  volume  = {40},
  number  = {9},
  pages   = {1887--1902},
  doi     = {10.1016/j.jbiomech.2006.09.003},
  pmid    = {17254589},
  issn    = {0021-9290}
}

@article{b2,
  author  = {Taylor, C. A. and Figueroa, C. A.},
  title   = {Patient-specific modeling of cardiovascular mechanics},
  journal = {Annual Review of Biomedical Engineering},
  year    = {2009},
  volume  = {11},
  pages   = {109--134},
  doi     = {10.1146/annurev.bioeng.10.061807.160521},
  pmid    = {19400706},
  issn    = {1523-9829}
}

@article{b3,
  author  = {Wilson, Nathan M. and Ortiz, Ana K. and Johnson, Allison B.},
  title   = {The Vascular Model Repository: A Public Resource of Medical Imaging Data and Blood Flow Simulation Results},
  journal = {Journal of Medical Devices},
  year    = {2013},
  volume  = {7},
  number  = {4},
  pages   = {040932},
  doi     = {10.1115/1.4025983},
  pmid    = {24895523},
  issn    = {1932-6181}
}

@article{b4,
  author  = {Di Martino, E. S. and Guadagni, G. and Fumero, A. and Ballerini, G. and Spirito, R. and Biglioli, P. and Redaelli, A.},
  title   = {Fluid-structure interaction within realistic three-dimensional models of the aneurysmatic aorta as a guidance to assess the risk of rupture of the aneurysm},
  journal = {Medical Engineering \& Physics},
  year    = {2001},
  volume  = {23},
  number  = {9},
  pages   = {647--655},
  doi     = {10.1016/s1350-4533(01)00093-5},
  pmid    = {11755809},
  issn    = {1350-4533}
}

@article{b5,
  author  = {Bianchi, Daniele and Monaldo, Elisabetta and Gizzi, Alessio and Marino, Michele and Filippi, Simonetta and Vairo, Giuseppe},
  title   = {A FSI computational framework for vascular physiopathology: A novel flow-tissue multiscale strategy},
  journal = {Medical Engineering \& Physics},
  year    = {2017},
  volume  = {47},
  pages   = {25--37},
  doi     = {10.1016/j.medengphy.2017.06.028},
  pmid    = {28690045},
  issn    = {1350-4533}
}

@article{b6,
author = {Roberts, David R. and Bahn, Volker and Ciuti, Simone and Boyce, Mark S. and Elith, Jane and Guillera-Arroita, Gurutzeta and Hauenstein, Severin and Lahoz-Monfort, José J. and Schröder, Boris and Thuiller, Wilfried and Warton, David I. and Wintle, Brendan A. and Hartig, Florian and Dormann, Carsten F.},
title = {Cross-validation strategies for data with temporal, spatial, hierarchical, or phylogenetic structure},
journal = {Ecography},
volume = {40},
number = {8},
pages = {913-929},
doi = {https://doi.org/10.1111/ecog.02881},
eprint = {https://nsojournals.onlinelibrary.wiley.com/doi/pdf/10.1111/ecog.02881},
year = {2017}
}

@article{b7,
  author  = {Shibeshi, Shewaferaw S. and Collins, William E.},
  title   = {The Rheology of Blood Flow in a Branched Arterial System},
  journal = {Applied Rheology},
  year    = {2005},
  volume  = {15},
  number  = {6},
  pages   = {398--405},
  doi     = {10.1901/jaba.2005.15-398},
  pmid    = {16932804},
  issn    = {1617-8106}
}

@article{b8,
  author  = {Pfaller, Martin R. and Pham, Jonathan and Wilson, Nathan M. and Parker, David W. and Marsden, Alison L.},
  title   = {On the Periodicity of Cardiovascular Fluid Dynamics Simulations},
  journal = {Annals of Biomedical Engineering},
  year    = {2021},
  volume  = {49},
  number  = {12},
  pages   = {3574--3592},
  doi     = {10.1007/s10439-021-02796-x},
  pmid    = {34169398},
  issn    = {0090-6964}
}

@article{b9,
  author  = {D. Lin and S. Kenjere\v{s}},
  title   = {Towards fast and reliable estimations of 3D pressure, velocity and wall shear stress in aortic blood flow: {CFD}-based machine learning approach},
  journal = {Computers in Biology and Medicine},
  volume  = {191},
  pages   = {110137},
  year    = {2025},
  doi     = {10.1016/j.compbiomed.2025.110137}
}

@article{b10,
  author    = {Chatpattanasiri, Chotirawee and Ninno, Federica and Stokes, Catriona and Dardik, Alan and Strosberg, David and Aboian, Edouard and von Tengg-Kobligk, Hendrik and D{\'i}az-Zuccarini, Vanessa and Balabani, Stavroula},
  title     = {{ML-ROM} wall shear stress prediction in patient-specific vascular pathologies under a limited clinical training data regime},
  journal   = {PLoS One},
  year      = {2025},
  volume    = {20},
  number    = {6},
  pages     = {e0325644},
  month     = jun,
  doi       = {10.1371/journal.pone.0325644},
  pmid      = {40504795},
  pmcid     = {PMC12161591},
  issn      = {1932-6203},
  publisher = {Public Library of Science}
}

@article{b11,
title = {A predictive surrogate model based on linear and nonlinear solution manifold reduction in cardiovascular FSI: A comparative study},
journal = {Computers in Biology and Medicine},
volume = {189},
pages = {109959},
year = {2025},
issn = {0010-4825},
doi = {https://doi.org/10.1016/j.compbiomed.2025.109959},
url = {https://www.sciencedirect.com/science/article/pii/S0010482525003105},
author = {M. {Barzegar Gerdroodbary} and Sajad Salavatidezfouli}
}

@article{b12,
  author    = {Rajhi, Wajdi and Ahmed, Zakarya and Ali, Ali B. M. and Alizadeh, As'ad and Hussein, Zahraa Abed and Sawaran Singh, Narinderjit Singh and Louhichi, Borhen and Aich, Walid},
  title     = {Use of proper orthogonal decomposition and machine learning for efficient blood flow prediction in cerebral saccular aneurysms},
  journal   = {Scientific Reports},
  year      = {2025},
  volume    = {15},
  number    = {1},
  pages     = {32335},
  month     = sep,
  doi       = {10.1038/s41598-025-17823-3},
  issn      = {2045-2322},
  url       = {https://doi.org/10.1038/s41598-025-17823-3}
}

@article{b13,
  author    = {Lannelongue, Vincent and Garnier, Paul and Jeken-Rico, Pablo and Goetz, Aur{\`e}le and Meliga, Philippe and Chau, Yves and Hachem, Elie},
  title     = {Physics constrained graph neural network for real time prediction of intracranial aneurysm hemodynamics},
  journal   = {npj Digital Medicine},
  year      = {2026},
  volume    = {9},
  number    = {1},
  pages     = {212},
  month     = feb,
  doi       = {10.1038/s41746-026-02404-z},
  issn      = {2398-6352},
  url       = {https://doi.org/10.1038/s41746-026-02404-z}
}

@article{b14,
title = {Robust prediction of parameterized cardiovascular hemodynamics using deep operator networks with time normalization},
journal = {Computer Methods and Programs in Biomedicine},
volume = {280},
pages = {109311},
year = {2026},
issn = {0169-2607},
doi = {https://doi.org/10.1016/j.cmpb.2026.109311},
url = {https://www.sciencedirect.com/science/article/pii/S0169260726000799},
author = {Junki Hong and Bomi Lee and Adelle Ria Persad and Hyunwoo Song and Changhee Min and Jae-Hak Jeong and Alireza Doostan and Yong-Hwa Park}
}

@article{b15,
  author  = {Ryan, William and Taylor-LaPole, Alyssa and Olufsen, Mette S. and Vyshemirsky, Vladislav and Husmeier, Dirk},
  title   = {Physics-Informed Neural Operators for Parameter Inference in Multi-vessel Cardiovascular Networks},
  journal = {Annals of Biomedical Engineering},
  year    = {2026},
  month   = jun,
  doi     = {10.1007/s10439-026-04219-1},
  pmid    = {42323515},
  issn    = {0090-6964},
  note    = {Epub ahead of print}
}

@article{b16,
  author  = {Wu, Runxin and Ferreira, Guinevere and Khan, Nusrat Sadia and Mahmud, Samreen T. and Stoop, Jorik and Sohn, Lydia L. and Leopold, Jane A. and Randles, Amanda},
  title   = {Digital twins and digital models of the human circulatory system},
  journal = {Nature Reviews Bioengineering},
  year    = {2026},
  volume  = {4},
  number  = {7},
  pages   = {594--613},
  month   = jul,
  doi     = {10.1038/s44222-026-00427-5},
  issn    = {2731-6092},
  url     = {https://doi.org/10.1038/s44222-026-00427-5}
}

@article{b17,
  author  = {Osman, Deen and Sel, Kaan and Spatz, Erica and Jafari, Roozbeh},
  title   = {Cardiovascular digital twins using a Windkessel physics informed neural network},
  journal = {npj Digital Medicine},
  year    = {2026},
  volume  = {9},
  number  = {1},
  pages   = {443},
  month   = apr,
  doi     = {10.1038/s41746-026-02610-9},
  issn    = {2398-6352}
}

@article{b19,
  author  = {Coorey, Genevieve and Figtree, Gemma A. and Fletcher, David F. and Snelson, Victoria J. and Vernon, Stephen Thomas and Winlaw, David and Grieve, Stuart M. and McEwan, Alistair and Yang, Jean Yee Hwa and Qian, Pierre and O'Brien, Kieran and Orchard, Jessica and Kim, Jinman and Patel, Sanjay and Redfern, Julie},
  title   = {The health digital twin to tackle cardiovascular disease---a review of an emerging interdisciplinary field},
  journal = {npj Digital Medicine},
  year    = {2022},
  volume  = {5},
  number  = {1},
  pages   = {126},
  doi     = {10.1038/s41746-022-00640-7},
  issn    = {2398-6352}
}

@article{b20,
  author  = {Gasser, T. Christian and Ogden, Ray W. and Holzapfel, Gerhard A.},
  title   = {Hyperelastic modelling of arterial layers with distributed collagen fibre orientations},
  journal = {Journal of the Royal Society Interface},
  year    = {2006},
  volume  = {3},
  number  = {6},
  pages   = {15--35},
  doi     = {10.1098/rsif.2005.0073},
  pmid    = {16849214},
  issn    = {1742-5689}
}

@article{b21,
  author  = {Doyle, Barry J. and Callanan, Anthony and McGloughlin, Timothy M.},
  title   = {A comparison of modelling techniques for computing wall stress in abdominal aortic aneurysms},
  journal = {Biomedical Engineering Online},
  year    = {2007},
  volume  = {6},
  pages   = {38},
  doi     = {10.1186/1475-925X-6-38},
  pmid    = {17949494},
  issn    = {1475-925X}
}

@article{b22,
   title={Machine Learning for Fluid Mechanics},
   volume={52},
   ISSN={1545-4479},
   url={http://dx.doi.org/10.1146/annurev-fluid-010719-060214},
   DOI={10.1146/annurev-fluid-010719-060214},
   number={1},
   journal={Annual Review of Fluid Mechanics},
   publisher={Annual Reviews},
   author={Brunton, Steven L. and Noack, Bernd R. and Koumoutsakos, Petros},
   year={2020},
   month=Jan, pages={477–508} 
   }

@inproceedings{b23,
author = {Ke, Guolin and Meng, Qi and Finley, Thomas and Wang, Taifeng and Chen, Wei and Ma, Weidong and Ye, Qiwei and Liu, Tie-Yan},
title = {LightGBM: a highly efficient gradient boosting decision tree},
year = {2017},
isbn = {9781510860964},
publisher = {Curran Associates Inc.},
address = {Red Hook, NY, USA},
booktitle = {Proceedings of the 31st International Conference on Neural Information Processing Systems},
pages = {3149–3157},
numpages = {9},
location = {Long Beach, California, USA},
series = {NIPS'17}
}

@article{b24,
title = {Physics-informed neural networks: A deep learning framework for solving forward and inverse problems involving nonlinear partial differential equations},
journal = {Journal of Computational Physics},
volume = {378},
pages = {686-707},
year = {2019},
issn = {0021-9991},
author = {M. Raissi and P. Perdikaris and G.E. Karniadakis}
}

\end{document}